\documentclass[11pt,a4paper]{article}
\usepackage[hyperindex,breaklinks]{hyperref}
\usepackage[hyperref]{acl2019}
\usepackage{times}
\usepackage{latexsym}
\usepackage{color}
\usepackage{url}
\usepackage[utf8]{inputenc}
\usepackage[T1]{fontenc}
\usepackage{tabularx}
\usepackage{booktabs}
\usepackage{float}

\aclfinalcopy 
\def\aclpaperid{143} 

\newcommand{\pulrad}[1]{\raisebox{1.5ex}[0pt]{#1}}

\def\to{$\rightarrow$}
\def\Tref#1{Table~\ref{#1}}

\def\C#1{\texttt{#1}}

\def\I{\it}
\def\B{\bf}

\title{English-Czech Systems in WMT19: Document-Level Transformer}

\author{Martin Popel, Dominik Macháček, Michal Auersperger, Ondřej Bojar \and Pavel Pecina\\
  Charles University, Faculty of Mathematics and Physics, \\
  Institute of Formal and Applied Linguistics, \\
  Malostranské náměstí 25, 118 00 Prague, Czech Republic \\
  \texttt{\textit{surname}@ufal.mff.cuni.cz}
 \\}
 
\date{}

\begin{document}
\maketitle
\begin{abstract}
We describe our NMT systems submitted to the WMT19 shared task in English\to{}Czech news translation.
Our systems are based on the Transformer model
implemented in either Tensor2Tensor (T2T) or Marian framework.


We aimed at improving the adequacy and coherence of translated documents
 by enlarging the context of the source and target.
Instead of translating each sentence independently,
 we split the document into possibly overlapping multi-sentence segments.
In case of the T2T implementation,
 this ``document-level''-trained system achieves a $+0.6$ BLEU improvement ($p<0.05$)
 relative to the same system applied on isolated sentences.
To assess the potential effect document-level models might have on lexical coherence,
 we performed a semi-automatic analysis, which revealed only a few sentences improved in this aspect.
Thus, we cannot draw any conclusions from this week evidence.
\end{abstract}

\section{Introduction}
Neural machine translation has reached a point,
 where the quality of automatic translation measured on isolated sentences
 is similar on average to the quality of professional human translations.
\citet{hassan-et-al:2018} report achieving a ``human parity'' on Chinese\to{}English news translation.
\citet[p.~291]{wmt18} report that our last year's English\to{}Czech system \citep{cuni-transformer-2018}
 was evaluated as significantly better ($p<0.05$) than the human reference.
However, it has been shown \citep{laubli,toral}
 that evaluating the quality of translation of news articles
 on isolated sentences without the context of the whole document is not sufficient.
It can bias the evaluation results
 because systems that ignore the context are not penalized in the evaluation for these context-related errors;
 and vice versa: systems (or humans) that take the context into account may be unfairly penalized.
\citet{laubli} show that while the difference between human and machine translation in adequacy is not significant when evaluated on isolated sentences,
 it is significant (humans are better) when evaluated on whole documents.
This suggests that there are some inter-sentential phenomena
 where MT applied on isolated sentences is lacking. 
 
Since assessing the performance of document-level systems is one of the goals of WMT19 \citep{wmt19},
 we decided to build NMT systems
 trained for translation of longer segments than single sentences.
In this paper, we describe our five NMT systems submitted to WMT19 English\to{}Czech news translation task (see \Tref{tab:systems}).
They are based on the Transformer model \citep{vaswani-et-al:2017} and on our submission from WMT18 \citep{cuni-transformer-2018}.
Our new contributions are 
(i) adaptation of the baseline single-sentence models
    to translate multiple adjacent sentences in a document at once,
    so the Transformer can attend to inter-sentence relations
    and achieve better document-level translation quality,
    as was already showed to be effective by \citet{jean};
and (ii) reimplementation of our last year's submission in the Marian framework \cite{mariannmt}.


\begin{table*}[]\centering\small
\begin{tabularx}{\linewidth}{l X}\toprule
official name              & description  \\\midrule
CUNI DocTransformer T2T    & Document level trained Transformer in T2T.\\
CUNI DocTransformer Marian & Document level trained Transformer in Marian. \\
CUNI Transformer T2T 2019  & Same model as CUNI~DocTransformer~T2T, but applied on single sentences (i.e. with no cross-sentence context). \\
CUNI Transformer T2T 2018  & Same model as in the last year \cite{cuni-transformer-2018}. \\
CUNI Transformer Marian    & Reimplementation of the last year's model in Marian. \\\bottomrule
\end{tabularx}
 \caption{Brief descriptions of our WMT19 systems.
 In the rest of the paper, we omit the CUNI (Charles University) prefix for brevity.}
 \label{tab:systems}
\end{table*}

This paper is organized as follows:
In Section~\ref{sec:setup}, we describe our training data and its augmentation to overlapping multi-sentence sequences.
We describe also the hyper-parameters of our models in the two frameworks.
Section~\ref{sec:doc-level-systems} follows with a description of the document-level decoding strategies.
Section~\ref{sec:results} reports and discusses the results of automatic (BLEU) evaluation.

\section{Experimental Setup}\label{sec:setup}

\subsection{Data sources}
\begin{table}\centering\small
\begin{tabular}{@{}l@{$\!$}rrr@{}}\toprule
                            & sentence  & \multicolumn{2}{c}{\hskip1cm words (k)}\\
\pulrad{data set}           & pairs (k) & EN       & CS    \\\midrule
CzEng 1.7                   &  57~065   &  618~424 &   543~184 \\
Europarl v7                 &     647   &   15~625 &    13~000 \\
News Commentary v12         &     211   &    4~544 &     4~057 \\
CommonCrawl                 &     162   &    3~349 &     2~927 \\
WikiTitles                  &     361   &      896 &       840 \\    
EN NewsCrawl 2016--17       &  47~483   &  934~981 &           \\ 
CS NewsCrawl 2007--17       &  65~383   &          &   927~348 \\ 
CS NewsCrawl 2018           &  12~983   &          &   181~004 \\\midrule
total                       & 184~295   &1~577~819 & 1~672~360 \\\bottomrule
\end{tabular}
\caption{Training data sizes (in thousands).}
\label{tab:data}
\end{table}

Our training data (see \Tref{tab:data}) are constrained to the data allowed in the WMT2019 shared task.
``Transformer T2T 2018'' and ``Transformer Marian'' use only the data allowed in WMT2018, which does not include CS NewsCrawl 2018 and WikiTitles.
All the data were preprocessed, filtered and backtranslated by the same process as in \citet{cuni-transformer-2018}.
We selected the originally English part of newstest2016 for validation,
 following the idea of CZ/nonCZ tuning in \citet{cuni-transformer-2018},
 but excluding the CZ tuning because the WMT2019 test set was announced to contain
 only original English sentences and no translationese.

\subsection{Training Data Context Augmentation}\label{sec:augment}

In WMT19, all the training data from \Tref{tab:data}
 are available with document boundaries (and unlike in previous years the sentences are not shuffled).\footnote{
 In WikiTitles, each pair of titles is considered a separate document.
 We decided to upsample this source 23 times, but we have not evaluated the effect of this on the final quality.
}
We extracted all sequences of consecutive sentences with at most 1000 characters.\footnote{
 The limit of 1000 characters was chosen rather arbitrarily.
 A 1000-characters long sequence from our training data contains on average about 15 sentences (165 English and 144 Czech words).
}
Our context-augmented data consists of pairs of such sequences,
 where the source sequence has always the same number of sentences as the target sentence.
We separate the sentences in each sequence with a special token,\footnote{
 Any token not present in the training data can be used,
  but it should be included in the subword vocabulary.}
 so that we can easily extract sentence alignment after decoding.
We randomly shuffle the augmented training sequences,
 but we keep separately the authentic parallel and synthetic (backtranslated) data,
 so that we can apply \emph{concat backtranslation} \citep{cuni-transformer-2018}.

Note that this particular way of context augmentation implicitly upsamples
 sentences from longer documents relative to sentences from shorter documents.
We leave the analysis of this effect and possible alternative samplings for future work.

\subsection{Model Hyper-parameters}

\subsubsection{Tensor2Tensor} \label{sec:t2t-training}
Our three systems with ``T2T'' in the name are implemented in the Tensor2Tensor framework \citep{t2t-framework}, version 1.6.0.
The model and training parameters this year are identical to our last year's (WMT18) submission \citep{cuni-transformer-2018},
 with just two exceptions:
First, we trained on 10 GPUs instead of 8 GPUs, thus using the effective batch size of 29k subwords instead of 23k subwords.
Second, we used \C{max\_length=200} instead of 150.
This means we discard all training sequences longer than 200 subwords.
With our 32k joint subword vocabulary, a word contains on average 1.5 subwords.
Thus effectively, the sequence-length limit used in T2T training was in most cases lower than 1000 characters
 -- on average it was 785 characters.

\subsubsection{Marian}
Our two systems with ``Marian'' in the name use the Marian framework \citep{mariannmt}, in the latest stable version 1.7.6. 
We chose Marian for its fast and efficient training and decoding. 
Due to the good results of ``CUNI Transformer'' in WMT18 evaluation and lack of time and resources for exhaustive parameter search, we reconstructed all its hyperparameters in Marian wherever possible.
Therefore, we trained with the following options:

{\small
\begin{verbatim}
--type transformer --enc-depth 6 
--dec-depth 6 --dim-emb 1024
--transformer-dim-ffn 4096
--transformer-heads 16
--transformer-dropout 0.0
--transformer-dropout-attention 0.1
--transformer-dropout-ffn 0.1
--lr-warmup 20000
--lr-decay-inv-sqrt 20000
--optimizer-params 0.9 0.98 1e-09 
--clip-norm 5 --label-smoothing 0.1 
--learn-rate 0.0002
--exponential-smoothing
\end{verbatim}
}

We used the same learning rate as T2T and estimated the number of warmup training steps so the model consumed approximately the same number of sentences as T2T in warmup.
Instead of T2T's default SubwordTextEncoder, we used SentencePiece \citep{sentencepiece} with its default parameters to obtain a shared vocabulary of 32,000 entries from untokenized training data. 
We set the maximal sentence length to 150 and decoded with beam size 4.

We could not use Adafactor \citep{adafactor} optimizer as in T2T, because it is not implemented in Marian. We used Adam instead. 

We did not set the batch size manually, but used the \texttt{-{}-mini-\-batch-fit} parameter to determine the mini-batch size automatically based on sentence lengths to fit the available memory.
We estimated the workspace memory to 13,900 MB as the largest possible on our hardware.
We shuffled the training data before training and did not use any advanced reordering to fit more non-padding tokens into a training batch as in T2T. 

Another difference is the checkpoint averaging:
while our T2T models are (uniform) averages of the last 8 checkpoints from the last 8 hours of training,
 our Marian models use the exponential moving average regularization method (\texttt{-{}-expo\-nen\-tial-smoothing}) applied after each update, as suggested by the Marian authors.


\subsection{Training}

\begin{table}\centering\small
\begin{tabular}{@{}l@{$\!$}rrr@{}}\toprule
systems      & \#GPUs       & GPU memory & GPU type \\\midrule
T2T 2018     &  8  & 11GB & GTX 1080 Ti \\
T2T 2019     & 10  & 11GB & GTX 1080 Ti  \\
Marian       &  8  & 16GB & Quadro P5000  \\
\bottomrule
\end{tabular}
\caption{Hardware used for our systems.}\label{tab:hw}
\end{table}

The summary of hardware used for training is in \Tref{tab:hw}. First, we trained a non-document models on single sentences, on concatenation of out-domain authentic data and in-domain synthetic datasets. We trained ``Transformer Marian'' model for 17 days until the epoch 18. We observed the last improvement in validation BLEU at 15 days and 18 hours of training, in step 1,266M, which we selected as the final model ``Transformer Marian''.
The ``DocTransformer T2T'' model was trained for 9 days (660k steps).

\section{Document-Level Systems}\label{sec:doc-level-systems}

Our document-level models were created by training on the context-augmented data described in Section~\ref{sec:augment}.
We used different strategies for document-level decoding in Marian and in T2T.

\subsection{Decoding in Marian}

For``DocTransformer Marian'' decoding,
 we decided to reduce the context to up to three consecutive sentences
 because decoding of longer contexts was time-consuming and our time was constrained.
Each sentence appeared as the first, second or third sentence in a 3-sentence context
 (1st/3, 2nd/3, 3rd/3) if possible.\footnote{
  For the first sentence in a document only 1st/3 is possible,
   for the second sentence only 1st/3 or 2nd/3 is possible, etc.
 }
We experimented also with a 2-sentence context (1st/2, 2nd/2)
 and no context (1st/1, i.e. the baseline).
 
We compared dev-set BLEU scores of these six setups
 and selected the following strategy for the selection of the final translation:
For each sentence, if possible and if the translation is ``valid'', use 2nd/3.
If not possible or ``valid'', use 1st/3, followed by 2nd/2, 1st/2 and 1st/1.

We consider a translation ``valid''
 if it contains the same number of sentences 
 (delimited by a special sentence-boundary character) as the input.
We excluded translations containing a given word more than 20 times
 and translations with a word longer than 49 characters.
This rule detected non-meaningful outputs that we observed in validation.
We decided to not use 3rd/3 because these translations were the least accurate ones.

Based on the validation BLEU scores,
 we selected two checkpoints for the final document-level translation.
The checkpoint at 2,044M steps was used for 1st/3, 2nd/3 and 2nd/2.
The checkpoint at 1,775M steps was used elsewhere (1st/2 and 1st/1).


\subsection{Decoding in T2T}

In an initial experiment, we split the test set into non-overlapping sequences of sentences with at most 1000 characters,
 following the maximum sequence length used in training.
We realized that the translation quality is very low, especially close to the end of each translated sequence.
Sometimes the number of output sentences
 (detected based on the special separator character)
 was different than the number of input sentences.
We hypothesized that the reason of low quality is that there are not enough 1000-character sequences in the training data (cf. Section~\ref{sec:augment}).
With non-overlapping splits, we achieved the best dev-set BLEU,
 when lowering the limit to about 700 characters.

We further experimented with overlapping splits,
 where each sequence to be translated consists of
\begin{itemize}
\item pre-context: sentences which are ignored in the translation and serve only as a context for better translation of the main content,
\item main content: sentences which are used for the final translation,
\item post-context: sentences which are ignored, similarly to the pre-context.
\end{itemize}

Based on a small dev-set BLEU hyper-parameter search,
 we selected the following length limits:
 pre-context of up to 200 characters (splitting on word boundaries),
 main content of up to 500 characters (whole sentences only)
 and post-context of up to 900 characters minus the length of the pre-context and main content (whole sentences only).
After the main decoding,
 we joined together the translations of main contents of all sequences.
In rare cases (8 sentences out of 3611),
 when there were not enough sentences in the translated sequence,
 we used a single-sentence translation as a backup.

\subsection{Post-processing}

For T2T systems, we used the same post-processing as last year \cite{cuni-transformer-2018}:
 We deleted the repetitions of phrases of one to four words appearing directly after each other more than two times, and converted the quotation symbols to „lower and upper``.
This is considered as standard in Czech formal texts.
For Marian, we applied only the conversion of quotation symbols.

\begin{table}[t]\centering\small
 \begin{tabular}{l@{~~}r@{~~}r@{~~}r@{~~}}\toprule
                       & BLEU     &   BLEU  &   chrF2   \\
 system                & uncased  &   cased &   cased   \\ \midrule
 DocTransformer~T2T    & \B 31.03 &\B 29.94 & \I 0.5628 \\
 Transformer~T2T~2018  & \I 30.93 &\I 29.86 & \B 0.5630 \\\cmidrule(lr){2-3}
 Transformer~T2T~2019  &    30.42 &   29.39 &    0.5552 \\\cmidrule(lr){2-3}
 DocTransformer~Marian &    29.17 &   28.14 &    0.5466 \\
 Transformer~Marian    &    29.20 &   28.13 &    0.5474 \\
 UEdin                 &    29.00 &   27.89 &    0.5516 \\
 \bottomrule
 \end{tabular}
 \caption{Automatic 
  evaluation on \C{newstest2019}.
  Significantly different BLEU scores ($p<0.05$ bootstrap resampling)
   are separated by a horizontal line.
 }\label{tab:wmt19}
 \end{table}

\section{Results}\label{sec:results}

\subsection{Automatic Evaluation}

\Tref{tab:wmt19} reports the automatic metrics of our English\to{}Czech systems submitted to WMT2019, plus the best other system -- UEdin (Marian system trained by University of Edinburgh). 
The automatic metrics are calculated using sacreBLEU 1.3.2 \cite{post:2018} and their signatures are:
\begin{itemize}
\item BLEU+case.mixed+lang.en-cs+numrefs.1+smooth.exp+tok.13a,
\item BLEU+case.lc+lang.en-cs+numrefs.1+smooth.exp+tok.intl and
\item chrF2+case.mixed+lang.en-cs+numchars.6+numrefs.1+space.False.
\end{itemize}

\subsection{Explaining the Difference of T2T and Marian}

The two comparable systems using the closest possible settings we were able to achieve and identical data, ``Transformer Marian'' and ``Transformer T2T 2018'', did not perform equally. The last year's T2T system was around 1.73 BLEU better at the point, where both systems had enough training time to converge. We hypothesize this was caused by the parameters, in which they differ: (i) Marian uses Adam optimizer, T2T Adafactor; (ii) Marian had 8 16GB GPUs and T2T 8 11GB GPUs, it means 128GB vs 88GB in total. We assume Marian is not as effective in memory usage, or we used bigger than optimal memory (and thus batch) size; 
(iii) Marian uses different batch ordering; (iv) in Marian, we used the exponential moving average, T2T used uniform averaging of the last 8 checkpoints. 

\subsection{Doc-Level Evaluation}

We hypothesized that by providing the translation model with larger attendable context, the resulting translations display larger lexical consistency. We could demonstrate it by finding less examples where an English polysemous word is translated to two or more Czech non-synonymous lemmata within one document.

To evaluate the hypothesis, we word-aligned the source and target sentences using \texttt{fast\_align} \cite{dyer-etal-2013-simple}.\footnote{
 To improve the reliability of automatic word alignments, we trained them on the translations together with the first 500k sentences of CzEng 1.7.
 Only the intersection of the source-to-target and target-to-source alignments was considered.}
We then lemmatized the aligned words (both English and Czech) using MorphoDiTa \cite{strakova14} and considered all instances where a single English lemma was aligned to at least two Czech lemmata in a single document.
Since our focus was on evaluating the difference between non-context and document-level models, we selected only the English lemmata with different number of aligned Czech lemmata in the two types of systems.
Two pairs of models were compared: ``DocTransformer T2T'' vs. ``Transformer T2T 2019'' and ``DocTransformer Marian'' vs. ``Transformer Marian''.
The final pool of examples was evaluated manually.

We found only one and three instances for the Marian and T2T models, respectively, where the document-level variant performed better than the non-context variant. The examples are shown in Table~\ref{tab:doclevel-examples}. We also found a possible counter-example where the document-level model performed worse than the non-context model, but the evaluation is not clear-cut. The example is shown in Table~\ref{tab:doclevel-examples-neg}.

Because there are too few examples for any meaningful quantitative analysis, we conclude more data is needed to evaluate the potential benefit a document-level model could have on lexical consistency.
By doing the manual evaluation, we found the cases where the inter-sentential context is necessary for determining the correct meaning of a polysemous word are rare.

\section{Conclusion}
We were not able to replicate our last year's T2T system in Marian,
 but we acknowledge several differences in the setup.
We were not able to improve the sentence-level Marian system BLEU
 by adding a context of up to three sentences.
Our document-level trained T2T system achieved an insignificant  improvement ($+0.1$ BLEU) over our last year's sentence-level T2T system,
 but applying this system on sentences led to a significant worsening
 ($-0.6$ BLEU).

\section*{Acknowledgments}
This research was supported by the Czech Science Foundation (grant n.\ 19-26934X) and the Grant Agency of Charles University (grant n.\ 978119). The experiments were conducted using language resources distributed by the LINDAT/CLARIN project of the Ministry of Education, Youth and Sports of the Czech Republic (LM2015071).
 

\bibliography{biblio}
\bibliographystyle{acl_natbib}


\def\lemma#1{\textbf{#1}}
\def\transl#1{(\textit{#1})}
\def\context#1{\textit{#1}}
\begin{table*}
\section{Appendix}\label{sec:appendix}
~\\
\centering\small
 \begin{tabularx}{\linewidth}{@{}l X@{}}
 \toprule
 source & [...] to meet Craig Halkett's header across goal. The hosts were content to let Rangers play in front of them, knowing they could trouble the visitors at set pieces. And that was the manner in which the crucial \lemma{goal} came. Rangers conceded a free-kick [...] \\
 T2T & A to byl způsob, jakým přišel rozhodující \lemma{cíl} \transl{aim}. \\
 T2T-doc & A to byl způsob, jakým přišel rozhodující \lemma{gól} \transl{goal}. \\
 \midrule
 source & Elizabeth Warren Will Take "Hard Look" At Running For President in 2020, Massachusetts Senator Says Massachusetts Senator Elizabeth Warren said on Saturday she would take a "hard look" at running for president following the midterm elections. During a town hall in Holyoke, Massachusetts, Warren confirmed she'd consider \lemma{running}. "It's time for women to go to Washington and fix our broken government and that includes a woman at the top," she said, according to The Hill. [...] \\
 T2T & Na radnici v Holyoke v Massachusetts Warrenová potvrdila, že uvažuje o \lemma{útěku} \transl{escape}. \\
 T2T-doc & Na radnici v Holyoke ve státě Massachusetts Warrenová potvrdila, že o \lemma{kandidatuře} \transl{candidacy} uvažuje. \\
 \midrule
 source & At 6am, just as Gegard Mousasi and Rory MacDonald were preparing to face each other, viewers in the UK were left stunned when the coverage changed to Peppa Pig. Some were unimpressed after they had stayed awake until the early hours especially for the \lemma{fight}. [...] \\
 T2T & Na některé to neudělalo žádný dojem, když zůstali vzhůru až do časných ranních hodin, zvláště kvůli \lemma{rvačce} \transl{crawl}. \\
 T2T-doc & Na některé to neudělalo žádný dojem, když zůstali vzhůru až do ranních hodin, zejména kvůli \lemma{zápasu} \transl{match}. \\
 \midrule

 source & [...] she felt "terrified of retaliation" and was worried about "being publicly humiliated." The 34-year-old says she is now seeking to overturn the \lemma{settlement} as she continues to be traumatized by the alleged incident. [...] \\ 
 Marian & Čtyřiatřicetiletá žena tvrdí, že se nyní snaží o zrušení \lemma{osady} \transl{village}, protože je nadále traumatizována \\ & údajným incidentem. \\
 Marian-doc & 34letá žena tvrdí, že nyní usiluje o zrušení \lemma{vyrovnání} \transl{compensation}, protože je nadále traumatizována údajným  incidentem. \\
 
 \bottomrule
 \end{tabularx}
 \caption{Examples of non-context model errors corrected by the document-level models.}
\label{tab:doclevel-examples}
\end{table*}

\begin{table*}[h]
\centering\small
 \begin{tabularx}{\linewidth}{@{}l X@{}}
 \toprule
 source & New cancer vaccine can teach the immune system to 'see' rogue cells New cancer vaccine can teach the immune system to 'see' rogue cells and kill them Vaccine teaches immune system to recognise rogue cells as part of treatment Method involves extracting immune cells from a \lemma{patient}, altering them in lab They can then 'see' a protein common to many cancers and then reinjected A trial vaccine is showing promising results in \lemma{patients} with a range of cancers. One woman treated with the vaccine, which teaches the immune system to recognise rogue cells, saw her ovarian cancer disappear for more than 18 months. The method involves extracting immune cells from a \lemma{patient}, altering them in the laboratory so they can "see" a protein common to many cancers called HER2, and then reinjecting the cells. \\
T2T & Nová protinádorová vakcína může naučit imunitní systém „vidět“ zlovolné buňky Nová protinádorová vakcína může naučit imunitní systém „vidět“ zlovolné buňky a zabít je. Vakcína učí imunitní systém rozpoznávat zlovolné buňky jako součást léčby Metoda zahrnuje odebrání imunitních buněk z \lemma{pacienta} a jejich změnu v laboratoři. Pak mohou vidět protein, který je společný mnoha nádorům, a znovu ho vstříknout. Zkušební vakcína vykazuje slibné výsledky u \lemma{pacientů} s řadou nádorových onemocnění. Jedna žena léčená vakcínou, která učí imunitní systém rozeznávat zlovolné buňky, byla svědkem vymizení rakoviny vaječníků na více než 18 měsíců. Metoda spočívá v odebrání imunitních buněk z \lemma{pacienta}, jejich přeměně v laboratoři, aby mohli „vidět“ protein, který je společný mnoha nádorům nazývaným HER2, a poté reinjekci buněk. \\
T2T-doc & Nová protinádorová vakcína může naučit imunitní systém „vidět“ zlovolné buňky Nová protinádorová vakcína může naučit imunitní systém „vidět“ zlovolné buňky a zabít je Vakcína učí imunitní systém rozpoznávat zlovolné buňky jako součást léčby Metoda zahrnuje extrakci imunitních buněk z \lemma{pacienta}, jejich změnu v laboratoři Poté mohou „vidět“ bílkovinu společnou mnoha nádorovým onemocněním a poté ji znovu nasadit Zkušební vakcína vykazuje slibné výsledky u \lemma{pacientů} s řadou nádorových onemocnění. Jedna žena léčená touto vakcínou, která učí imunitní systém rozpoznávat zlovolné buňky, byla svědkem vymizení rakoviny vaječníků na více než 18 měsíců. Tato metoda zahrnuje odebrání imunitních buněk od \lemma{pacientky} \transl{female patient}, jejich změnu v laboratoři, aby mohly „vidět“ bílkovinu, která je společná mnoha nádorům nazývaným HER2, a poté reinjekci buněk. \\
 \bottomrule
 \end{tabularx}
 \caption{The example of an error introduced by a document-level model.}
\label{tab:doclevel-examples-neg}
\end{table*}


\end{document}